# Diffusion denoiser-aided gyrocompassing

Gershon Ben-Arie [a,*,1], Daniel Engelsman [a], Rotem Dror [b], Itzik Klein [a]

[a] The Hatter Department of Marine Technologies, University of Haifa, Israel
[b] The Department of Information Systems, University of Haifa, Israel

ARTICLE INFO

ABSTRACT

An accurate initial heading angle is essential for efficient and safe navigation across diverse domains. Unlike magnetometers, gyroscopes can provide accurate heading reference independent of the magnetic disturbances in a process known as gyrocompassing. Yet, accurate and timely gyrocompassing, using low-cost gyroscopes, remains a significant challenge in scenarios where external navigation aids are unavailable. Such challenges are commonly addressed in real-world applications such as autonomous vehicles, where size, weight, and power limitations restrict sensor quality, and noisy measurements severely degrade gyrocompassing performance. To cope with this challenge, we propose a novel diffusion denoiser-aided gyrocompassing approach. It integrates a diffusion-based denoising framework with an enhanced learning-based heading extraction model. The diffusion denoiser processes raw inertial sensor signals before input to the deep learning model, performing the heading extraction. Our architecture uniquely combines probabilistic denoising and bidirectional temporal modeling, enabling sensor-agnostic noise reduction and sensor-specific orientation estimation, resulting in gyrocompassing with improved accuracy given the same time frame. Experiments using both simulated and real sensor data demonstrate that our proposed approach improves gyrocompassing accuracy, in terms of the cyclic root mean square error, by 26% compared to model-based gyrocompassing and by 15% compared to other learning-driven approaches, based on the same gyrocompassing duration. This advancement holds particular significance for ensuring accurate and robust navigation in autonomous platforms that incorporate low-cost gyroscopes within their navigation systems.

## 1. Introduction

The low cost and reduced size, weight, and power (SWaP) of micro-electromechanical systems (MEMS) technology drive the necessity for exploring their wider use in navigation, particularly for autonomous small vehicles in various applications [1]. Navigation solutions, providing position, velocity, and orientation, require initial conditions. Inertial sensors offer a means to determine the initial orientation [2,3], highlighting the importance of acquiring these initial angles promptly and with sufficient accuracy for effective navigation.

Specifically, the roll and pitch angles can be determined using closed-form analytical expressions based on the gravity components sensed by accelerometers [4]. However, determining the initial heading angle presents a greater challenge. Most systems rely on external aids to resolve heading, since low-cost MEMS inertial sensors lack the sensitivity required to detect the Earth's rotation rate [5–7]. These external aids may include consumer-grade magnetometers, satellite navigation systems, radio frequency (RF)-based localization, or visual sensor fusion techniques [8–13]. While these methods are practical and effective in many environments [14], their performance often degrades or becomes unavailable in challenging environments, such as underwater scenarios, GNSS-denied regions, areas with strong magnetic or electronic interference, or visually homogeneous environments where visual cues are limited or absent.

With advancements in MEMS technology, commercial off-the-shelf gyroscopes have begun to reach sensitivities approaching the Earth's rotation rate. Nevertheless, inherent bias and noise in MEMS sensors, sensitive to environmental conditions as vibrations and temperature [2, 15–18],continue to pose significant obstacles to achieving timely and accurate MEMS-based gyrocompassing.

To address these limitations, machine learning (ML) and deep learning (DL) approaches have been increasingly adopted in navigation tasks [19–26]. These methods have shown promise in identifying MEMS bias, mitigating sensor noise, and partially replacing traditional model-based alignment strategies. However, most existing approaches focus on inertial navigation, positioning, or calibration tasks rather

* Corresponding author.
*E-mail address:* gbenarie@campus.haifa.ac.il (G. Ben-Arie).
1 G. B. A. was supported by the Maurice Hatter Foundation.

https://doi.org/10.1016/j.measurement.2026.122370
Received 18 November 2025; Received in revised form 24 June 2026; Accepted 25 June 2026
Available online 8 July 2026

than on direct recovery of weak inertial signals required for gyrocompassing. The use of DL approaches for gyrocompassing was first proposed in [27] for stationary gyroscopes and elaborated in [28] to compensate for ocean currents and other disturbances. These works demonstrated more than 20% improvement in heading accuracy, depending on the gyrocompassing duration, but still operate directly on noisy sensor measurements without explicitly addressing signal denoising.

Recently, a new class of generative deep learning models has demonstrated the ability not only to synthesize realistic data but also to effectively denoise raw sensor signals. Among them, diffusion models have gained particular attention for their iterative noise estimation and removal process, originally developed for image denoising and generation tasks. More recently, diffusion-based approaches have been extended to time-series domains, including the synthesis of accelerometer and gyroscope data. However, most existing diffusion-based works focus on generic denoising, synthetic time-series generation, or IMU bias estimation, rather than on recovering physically meaningful inertial signals buried below the sensor noise floor. In particular, the use of diffusion models for noise reduction in navigation systems employing low-grade gyroscopes remains largely unexplored [29–34].

Recent advances in sequence-learning architectures have further expanded the range of deep-learning tools available for inertial sensing and navigation applications. Transformer-based models have demonstrated strong capability in capturing long-range temporal dependencies through self-attention mechanisms and have recently been applied to inertial odometry and localization problems [35,36], while temporal convolutional networks (TCNs) and related dilated convolution architectures have shown competitive performance for IMU denoising, error compensation, and sequence modeling tasks [25,37]. In parallel, self-supervised learning approaches have emerged as an attractive framework for inertial sensor processing by exploiting large quantities of unlabeled IMU data to learn robust motion representations and improve sensor signal quality [38].

While these approaches have demonstrated promising performance in sequence modeling, representation learning, and signal restoration tasks, the primary challenge addressed in the present work is the recovery of an extremely weak Earth-rotation-induced signal embedded within sensor noise. Unlike sequence-learning architectures that primarily focus on extracting temporal representations, diffusion models explicitly formulate denoising as an iterative noise estimation and removal process. This characteristic makes them particularly attractive for standalone MEMS gyrocompassing, where the useful signal may lie near or below the nominal sensor noise floor. Consequently, diffusion-based methods offer a fundamentally different perspective by directly targeting noise estimation and suppression rather than solely learning temporal feature representations.

Motivated by the potential of learning-based gyrocompassing approaches [27,28], the extremely low signal-to-noise ratio characteristics of MEMS-based gyrocompassing, and the potential ability of diffusion models to iteratively estimate and remove noise from such signals, we propose a stationary diffusion denoiser-aided gyrocompassing framework to provide a robust and accurate heading angle. Our main contributions are:

(i) **Data-Driven Physics-Based Diffusion Denoising:** We propose a diffusion-based denoising model trained on physics-based synthetic inertial data to recover extremely low-amplitude signals and improve standalone MEMS-based gyrocompassing accuracy.
(ii) **Synthetic Data Diffusion Model Training:** The diffusion model is trained using synthetic, physics-based data, offering greater flexibility in developing diffusion-based methods when real sensor data is not available.
(iii) **Robust Heading Extraction:** We introduce improvements to a baseline heading extraction model aimed at stabilizing training convergence and improving accuracy, robustness, and generalization performance.
(iv) **Robust Initialization Framework:** We present a learning-based heading initialization pipeline designed for reliable operation in GNSS-denied or magnetically disturbed environments.

The rest of the paper is organized as follows: Section 2 provides the problem formulation. Section 3 details the methodology used in this research. Section 4 presents the experimental results and analysis. Finally, Section 5 concludes the paper and discusses potential directions for future work.

## 2. Problem formulation

This section outlines the traditional gyrocompassing approach followed by a description of a recent deep learning-assisted gyrocompassing method.

### 2.1. Gyrocompassing

After obtaining the roll and pitch angles from the accelerometers readings, the heading angle can be estimated using the gyroscopes angular rates ($\omega_{ib}^{b}$) and the connection between them and earth rotation rate ($\omega_{ie}^{e}$):

$$\omega_{ib}^{b} = T_{n}^{b} T_{e}^{n} \omega_{ie}^{e} \tag{1}$$

where $T_{n}^{b}$ is the transformation matrix used to describe rotation between the navigation frame (n) to the body frame (b).

This transformation matrix is defined by [2]:

$$T_{n}^{b} = \begin{bmatrix} c_\theta c_\psi & c_\theta s_\psi & -s_\theta \\ s_\phi s_\theta c_\psi - c_\phi s_\psi & s_\phi s_\theta s_\psi + c_\phi c_\psi & s_\phi c_\theta \\ c_\phi s_\theta c_\psi + s_\phi s_\psi & c_\phi s_\theta s_\psi - s_\phi c_\psi & c_\phi c_\theta \end{bmatrix} \tag{2}$$

where $c$ is cosine, $s$ is sine, $\psi$ is yaw angle, $\theta$ is pitch angle, and $\phi$ is roll angle. The rotation matrix from earth centered earth fixed (ECEF) frame (e-frame) to the n-frame defined by:

$$T_{e}^{n} = \begin{bmatrix} -s_\varphi c_\lambda & -s_\varphi s_\lambda & c_\varphi \\ -s_\lambda & c_\lambda & 0 \\ -c_\varphi c_\lambda & -c_\varphi s_\lambda & -s_\varphi \end{bmatrix} \tag{3}$$

where $\lambda$ is the longitude and $\varphi$ is the latitude. Combining (1)–(3), the heading angle is obtained following [15]:

$$\mathrm{s}_\psi = -\omega_{ib,y}^{b} \mathrm{c}_\phi + \omega_{ib,z}^{b} \mathrm{s}_\phi \tag{4}$$

$$\mathrm{c}_\psi = \omega_{ib,x}^{b} \mathrm{c}_\theta + \omega_{ib,y}^{b} \mathrm{s}_\phi \mathrm{s}_\theta + \omega_{ib,z}^{b} \mathrm{c}_\phi \mathrm{s}_\theta \tag{5}$$

where, $\omega_{ib,x}^{b}$, $\omega_{ib,y}^{b}$, and $\omega_{ib,z}^{b}$ are the gyroscopes angular rate components in body frame, and then $\psi$ is solvable by [2]:

$$\psi = \mathrm{atan}_2(\mathrm{s}_\psi , \mathrm{c}_\psi) \tag{6}$$

In leveled conditions (6) reduces to:

$$\psi = \mathrm{atan}_2\left(-\omega_{ib,y}^{b}, \omega_{ib,x}^{b}\right) \tag{7}$$

To demonstrate the influence of latitude on the heading (7), we introduce (2) and (3) into (1) yielding:

$$\omega_{ib}^{b} = \omega_{ie}^{e} \begin{bmatrix} c_\psi c_\varphi \\ -s_\psi c_\varphi \\ -s_\varphi \end{bmatrix} \tag{8}$$

Using (8), we can assess the required sensitivity of the gyroscopes through $\omega_{ib,y}^{b}$ and $\omega_{ib,x}^{b}$. Utilizing (8), the maximum magnitude of these readings is:

$$||\omega_{ib,x}^{b}, \omega_{ib,y}^{b}|| = \omega_{ie} c\varphi \tag{9}$$

Thus, if gyrocompassing is performed at $\varphi = 32.11\,^\circ$ N, the maximal sensed signal affecting the heading calculation would be 0.0035 [°/s]. This means that latitude above or below the equator, contributes to the decay in the signal we are utilizing for gyrocompassing (16% in that case).

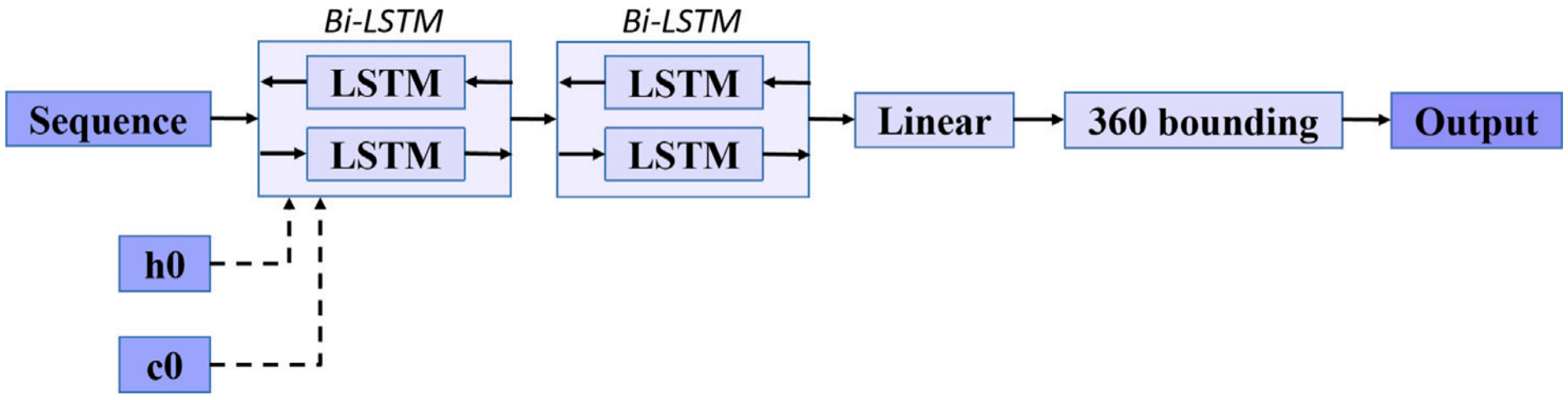

**Fig. 1.** Baseline Bi-Directional LSTM architecture.

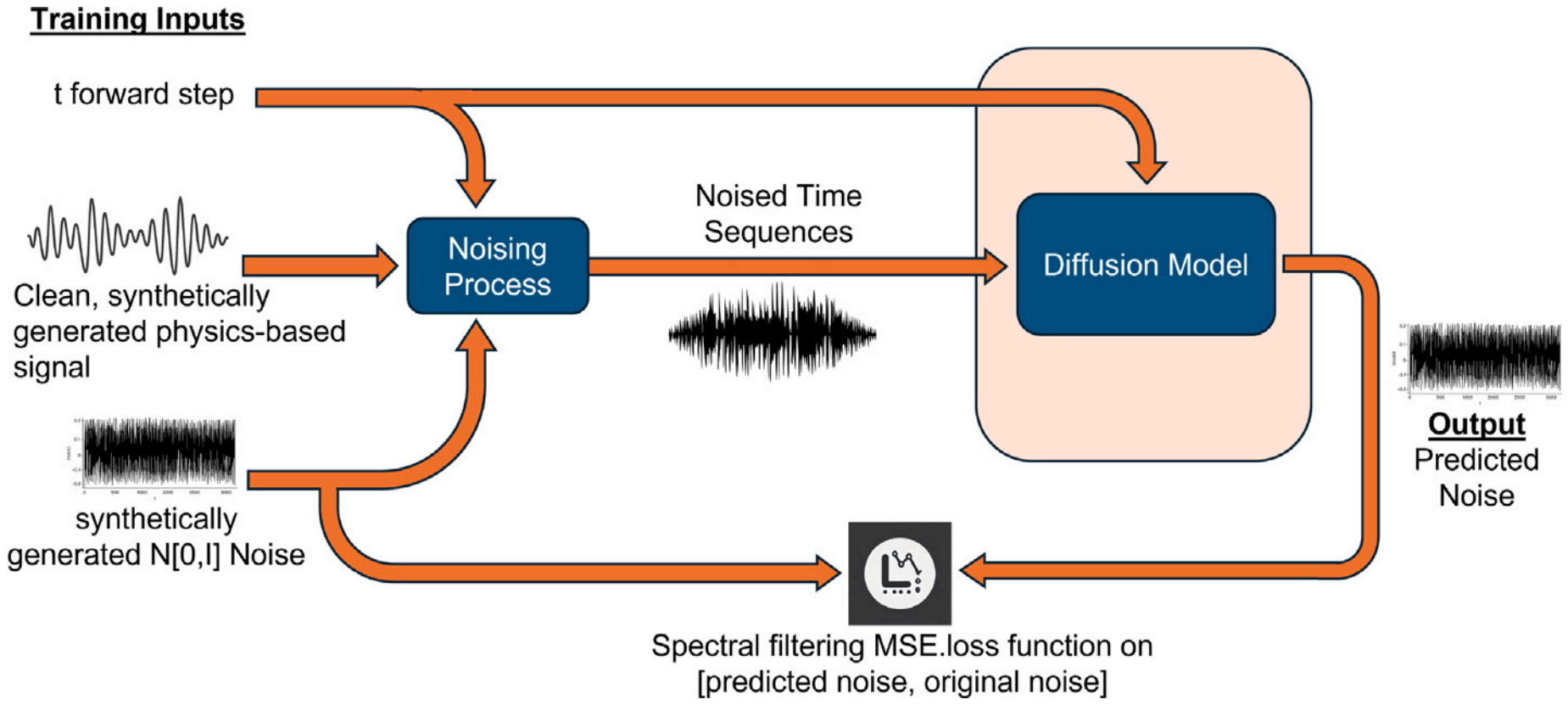

**Fig. 2.** Workflow of the proposed system's first training stage: noise-inference component training.

### 2.2. DL gyrocompassing

For a stationary, leveled, and calibrated platform, Engelsman and Klein [27], were the first to present a DL approach for gyrocompassing. The network architecture is based on a bi-directional long-short-term memory (LSTM) layers that process the gyroscope time sequence measurements and a fully connected (FC) layer is then connected to the LSTM output and predicts the platform heading angle bounded within a 0–360 degrees range. The bi-directional LSTM has an inner architecture of 2 LSTM layers and 24 hidden states, as illustrated in Fig. 1. This architecture encompasses 140581 trainable parameters.

## 3. Proposed approach

This section presents our proposed approach, following the same assumptions made in [27] of performing gyrocompassing on a stationary, leveled, and calibrated platform. We follow a top-down methodology, where we first explain the overall end-to-end system and then provide details on each model, including architecture, training methodology, hyperparameters, and data processing.

### 3.1. Proposed end-to-end system

To achieve enhanced heading angle accuracy and based on sensors noise consisting of both white and colored components [1,15,39,40],we designed a generic denoising model before the sensor specific heading extraction model for mitigating the white-noise component of the sensor measurements. The denoising model is based on diffusion model [32,41], and the heading extraction model is based on a bi-directional LSTM neural network. During training, we adopt the cyclic loss function as suggested in [27]. The diffusion model receives a noisy time sequence and predicts the noise induced on the signal. It iteratively removes the noise in steps until the signal is cleaned. The cleaned time sequence is then passed to the heading extraction model with the predicted heading as output.

The training process is twofold. In the first stage, the denoiser model is trained independently on synthetically generated, physics-based data augmented with white noise to determine its optimal parameters, as illustrated in Fig. 2, and in line with the diffusion model role as a generic white noise denoiser. In the second stage, the pre-trained denoiser is used to filter real gyroscope recordings, which serve as input for training the heading-extraction model, as shown in Fig. 3. After completing these two stages, a fully trained dual-model system is obtained, enabling heading-angle extraction and overall performance evaluation of the integrated system, as depicted in Fig. 4.

Two distinct datasets are therefore employed: synthetic data for training the denoiser, and real sensor data for training the heading-extraction model and for system-level assessment. The proposed architecture is designed to be applicable across different gyroscope sensor types: the denoiser training process is not tied to a specific sensor, while the heading-extraction model maintains a fixed architecture and requires only weight tuning based on recordings from the specific sensor type.

### 3.2. Diffusion denoiser

Our denoising approach is based on diffusion models. Diffusion models are probabilistic generative models that can generate data by iteratively denoising a sample that is pure noise. The process is based on the addition of a Gaussian noise to the training (source) data, thereby "corrupting" the original input in what is called a forward pass, where the goal of the model is to identify what noise was added to the source sample and restore the original data in a denoising process (backwards pass) [30,31]. The forward pass is a Markov process in which Gaussian noise is progressively added to data over t steps as follows:

$$q(x_t \mid x_{t-1}) \sim \mathcal{N}\left(x_t; \sqrt{1-\beta_t}\, x_{t-1}, \beta_t\, \mathbf{I}\right) \tag{10}$$

where $t = 1, \dots, T$, $x_t$ is the data at noising step $t$ (i.e., the data noised t times), and $\beta_t \in (\beta_{\min}, \beta_{\max})$ is the variance schedule determining the added noise versus residual signal at each step. The stochastic marginal distribution of $x_t$ given $x_0$ in a variance-preserving diffusion process is:

$$q(x_t \mid x_0) \sim \mathcal{N}\left(x_t; \sqrt{\hat{\alpha}_t} x_0, (1-\hat{\alpha}_t)\mathbf{I}\right) \tag{11}$$

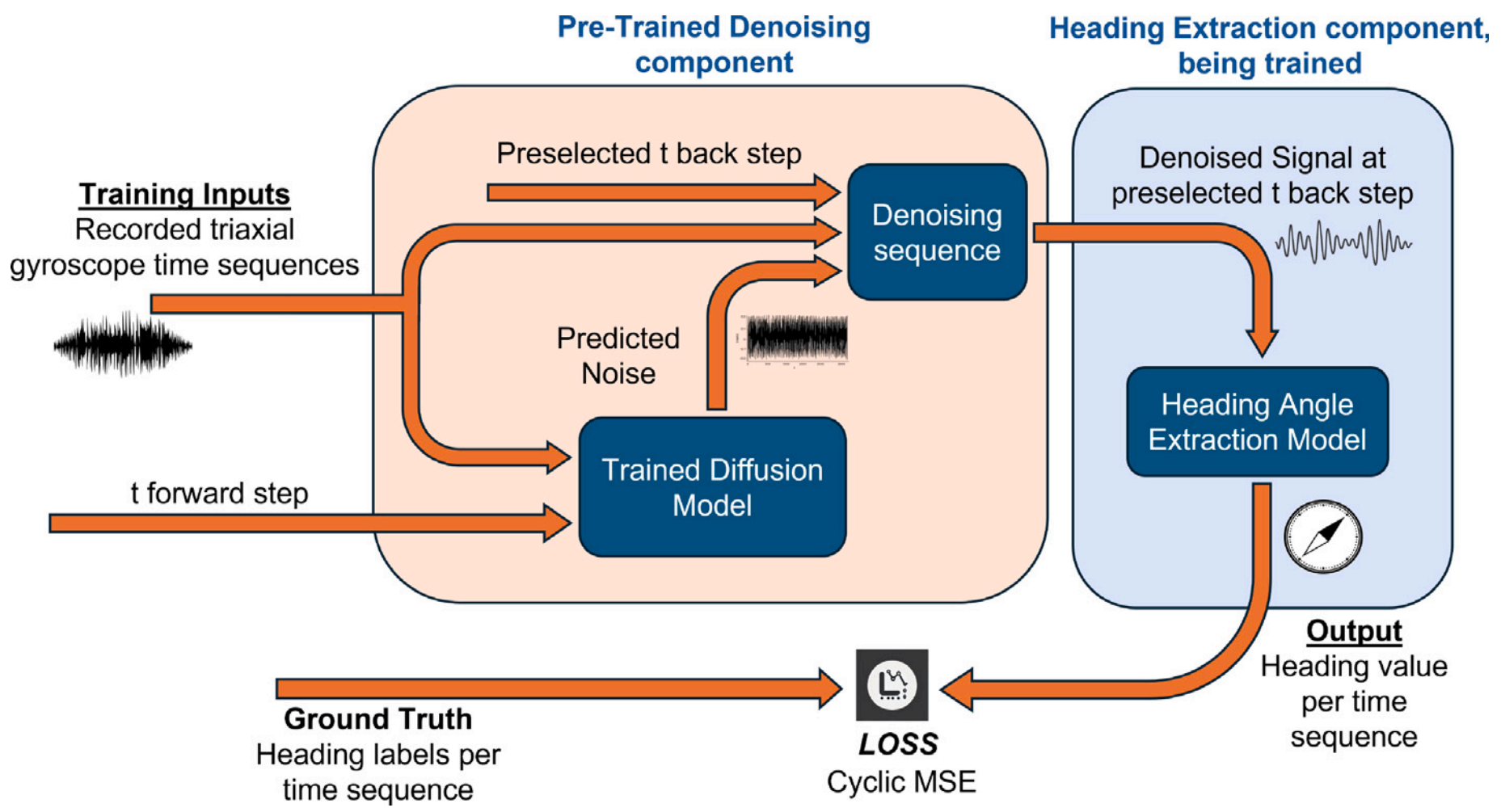

**Fig. 3.** Workflow of the proposed system's second training stage: heading-extraction model training.

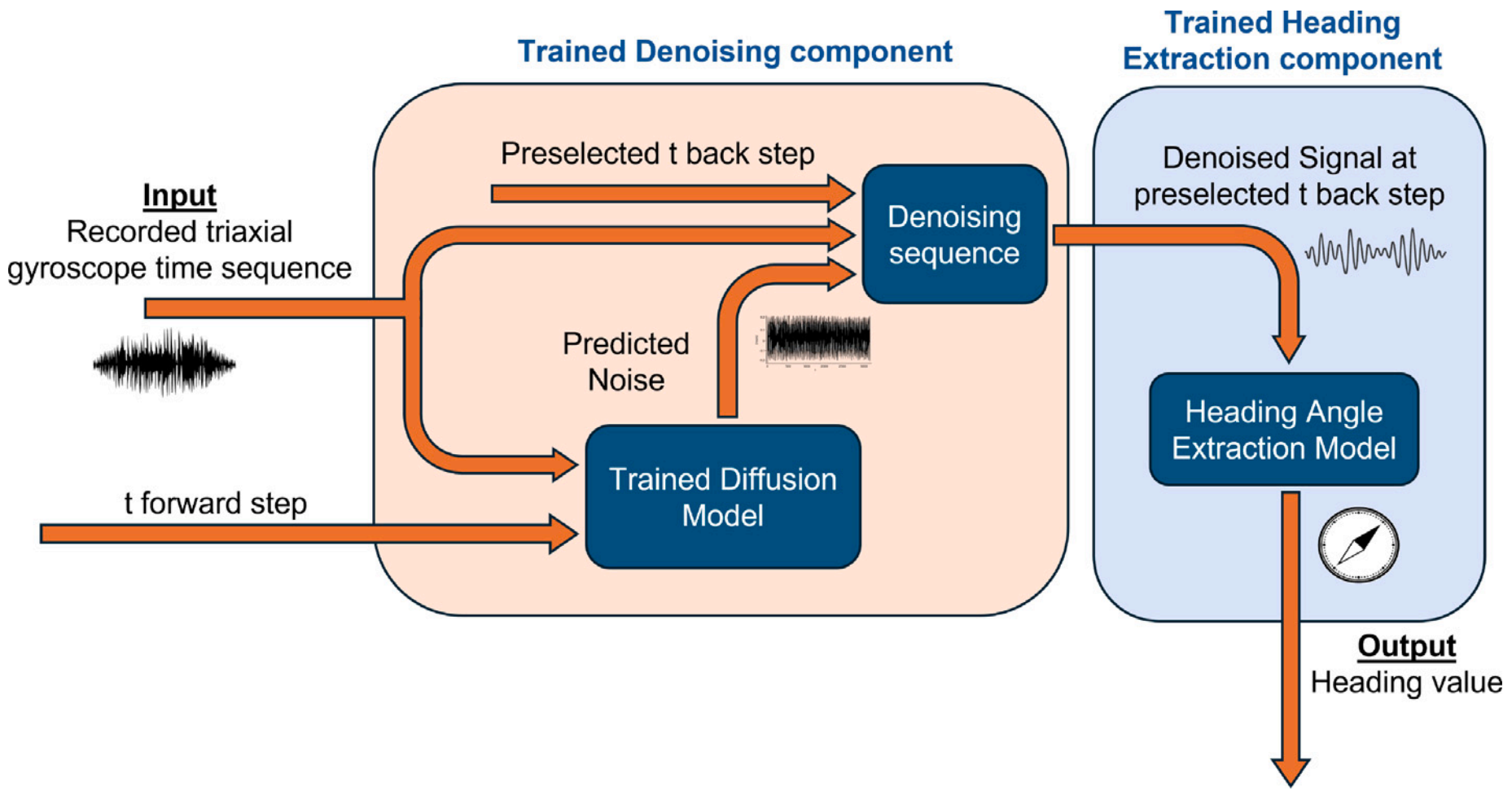

**Fig. 4.** Workflow of the proposed system's top-level architecture, illustrating the heading-angle extraction process performed by the trained models.

where:

$$\alpha_t = 1 - \beta_t \tag{12}$$

and $\hat{\alpha}_t$ is the cumulative product of $\alpha_t$:

$$\hat{\alpha}_t = \prod_{s=1}^{t} \alpha_s \tag{13}$$

The reverse process for denoising $x_t$ attempts to reverse the forward process by iteratively removing noise from $x_t$. This process is parameterized as:

$$p_\theta(x_{t-1} \mid x_t) \sim \mathcal{N}(x_{t-1}; \mu_\theta(x_t, t), \Sigma_\theta(x_t, t)) \tag{14}$$

where $\mu_\theta$ and $\Sigma_\theta$ are neural network outputs learned during training. The hyperparameters in the diffusion model are $T$, $\beta_{\min}$, and $\beta_{\max}$. $T$ is an hyperparameter that defines the number of maximal diffusion steps. A larger $T$ provides finer granularity for modeling the data distribution but increases computational cost. The variance schedule $\beta_t$ can be linearly or non-linearly spaced over $[\beta_{\min}, \beta_{\max}]$. A common choice, and the one we use is the linear, following:

$$\beta_t = \beta_{\min} + t \cdot \frac{\beta_{\max} - \beta_{\min}}{T} \tag{15}$$

where all the parameters in (15) are unitless. The selected values for the diffusion method in our implementation were: $T = 1000$, $\beta_{min} = 0.0001$ and $\beta_{max} = 0.0005$. This selection follows common practice in recent diffusion-based methods [32,34] for $T$ and, $\beta_{min}$. Regarding $\beta_{max}$, in contrast to generative diffusion models, the purpose of the current proposed model is denoising MEMS gyroscope signals, dominated by earth rotation, which guided selecting a relatively low $\beta_{max}$ for preserving the physical signal across diffusion steps and enabling effective recovery during the reverse process. Once calculated, $\beta_t$ is used to calculate $q(x_t \mid x_0)$ following (12), (13), and then (11) . These values were chosen based on real data analysis, signal-to-noise ratio evaluations, and empirical adjustments to match real-world conditions.

Finally, in practice, the noising process is based on using a clean signal $x_0$, predefined random noise $\epsilon$, $t_{step}$ noising iterations, $\beta_{\min}$, $\beta_{\max}$ parameters and then calculating $\beta_t$(15), $\alpha_t$ (12), and then $\hat{\alpha}_t$ (13). Finally, $x_t$ is defined according to:

$$x_t = \sqrt{\hat{\alpha}_t} x_0 + \sqrt{1 - \hat{\alpha}_t} \epsilon \tag{16}$$

At the heart of the denoising segment resides the diffusion model, predicting the noise added to the signal. Our diffusion model algorithm receives two inputs: time sequence, and $t_{step}$ which is the number of noising forward iterations. The model architecture consisted of two main functional components (Fig. 5): Insertion of the $t_{step}$ embeddings and noise identification.

The goal of the $t_{step}$ embeddings insertion component is to influence the bi-directional LSTM model's noise identification process by incorporating information about the amount of noise added. In our implementation, the embedding dimension was set to 20, as larger values (e.g., 100) were computationally expensive without significantly improving performance. Then, within the noise detection component, we used five bi-directional LSTM layers with 64 hidden states per layer to maximize the model's ability in capturing $t_{step}$ effects and noise patterns, while balancing computational cost. Consequently, the denoiser model contains 452,530 trainable parameters.

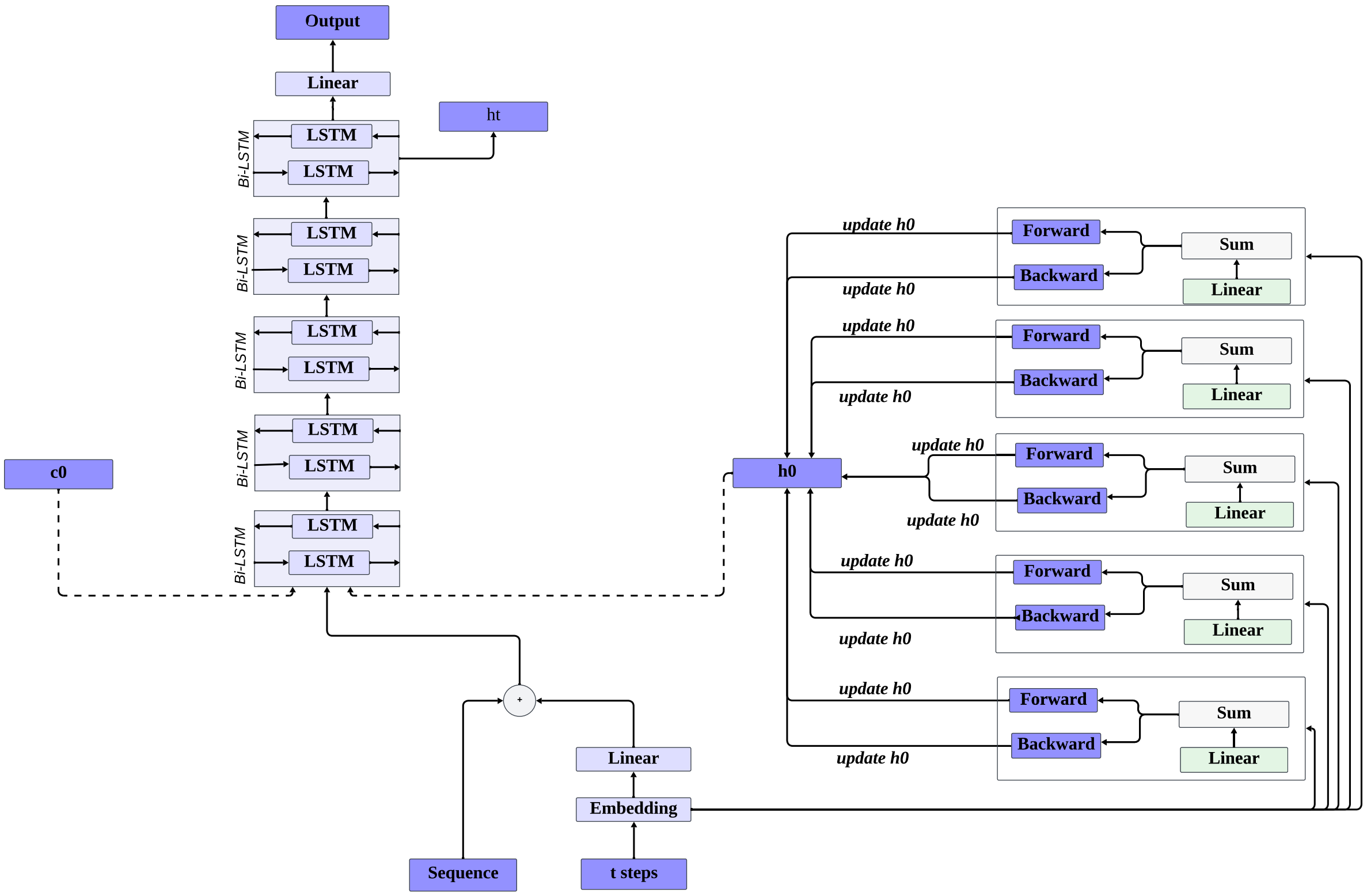

**Fig. 5.** Architecture of the diffusion model. A time sequence and a $t_{step}$ input are combined after embedding the $t_{step}$. The fused input passes through a bi-directional LSTM with $t_{step}$ embeddings added at each layer. The final output is generated via a fully connected layer.

As a loss function, we compared two alternatives suitable for our problem of supervised regression, mean squared error (MSE) and spectral filtering MSE. Although MSE is a commonly used loss function, our experiments showed that, when integrated with the downstream heading extraction model, it was not optimal for this application. Specifically, MSE penalizes all prediction errors uniformly rather than focusing on more dominant patterns. This is particularly challenging in our case due to the very low magnitude of the Earth-rotation signal relative to the sensor noise floor. Therefore, we adopted a spectral filtering MSE loss function [42,43], which suppresses the least dominant singular components of the inferred noise before comparing it to the ground-truth noise, thereby emphasizing energetically significant structures while reducing sensitivity to weak random fluctuations. In our preliminary tests, this approach showed an approximately 11% improvement in heading estimation accuracy over the standard MSE-based results. The implementation method we applied for the spectral filtering MSE used singular value decomposition (SVD) in the following manner:

$$M = USV^T \tag{17}$$

where $M$ is the matrix we want to apply the filter on, $U$ and $V$ are orthogonal matrices, and $S$ is a diagonal matrix containing values related to the importance of different frequency components in $M$.

Generally, small singular values correspond to frequencies that are less present in the data. Those are set to zero by applying a threshold-based approach:

$$S_{\text{filtered},i} = \begin{cases} S_i, & S_i > \tau S_{\max} \\ 0, & S_i \le \tau S_{\max} \end{cases} \tag{18}$$

where $S_{\text{filtered},i}$ denotes the $i$th filtered component of $S$, $S_i$ is the original $i$th component of the matrix $S$, $S_{\max}$ represents the maximum value among all $S_i$, and $\tau$ is the filtering threshold expressed as a ratio with respect to $S_{\max}$.

Then, after filtering, the matrix is reconstructed using the $U$, $S_{\text{filtered}}$, and $V$ components:

$$M_{\text{filtered}} = U S_{\text{filtered}} V^T \tag{19}$$

Given the reconstructed matrix, the loss is computed as the element-wise MSE between the filtered noise matrix predictions and the filtered noise matrix targets:

$$\mathcal{L}_{\text{SVD}} = \frac{1}{N}\sum_{i=1}^{N}(M_{\text{filtered},i} - M_{\text{target},i})^2 \tag{20}$$

where $N$ is the batch size, $M_{\text{filtered},i}$ is the filtered predicted noise matrix for the $i$th noised time sequence in the training batch, and $M_{\text{target},i}$ is the corresponding filtered ground truth noise matrix. After conducting several preliminary evaluations with different combinations of batch sizes and learning rates during the training phase, we determined that the optimal configuration, yielding the minimum loss, corresponds to a batch size of 50 time sequences, a learning rate of 0.001, and a filtering threshold $\tau$ of 0.1. This value is in line with the options examined in [43] and reflects the need to balance preservation of the low-magnitude Earth-rotation signal with suppression of medium and high-magnitude noise components. Accordingly, these parameters were adopted in conjunction with the ADAM optimizer for training the diffusion model.

Based on (16), the training process is carried on by iterations until convergence of predicted noise using the loss function (20), or until

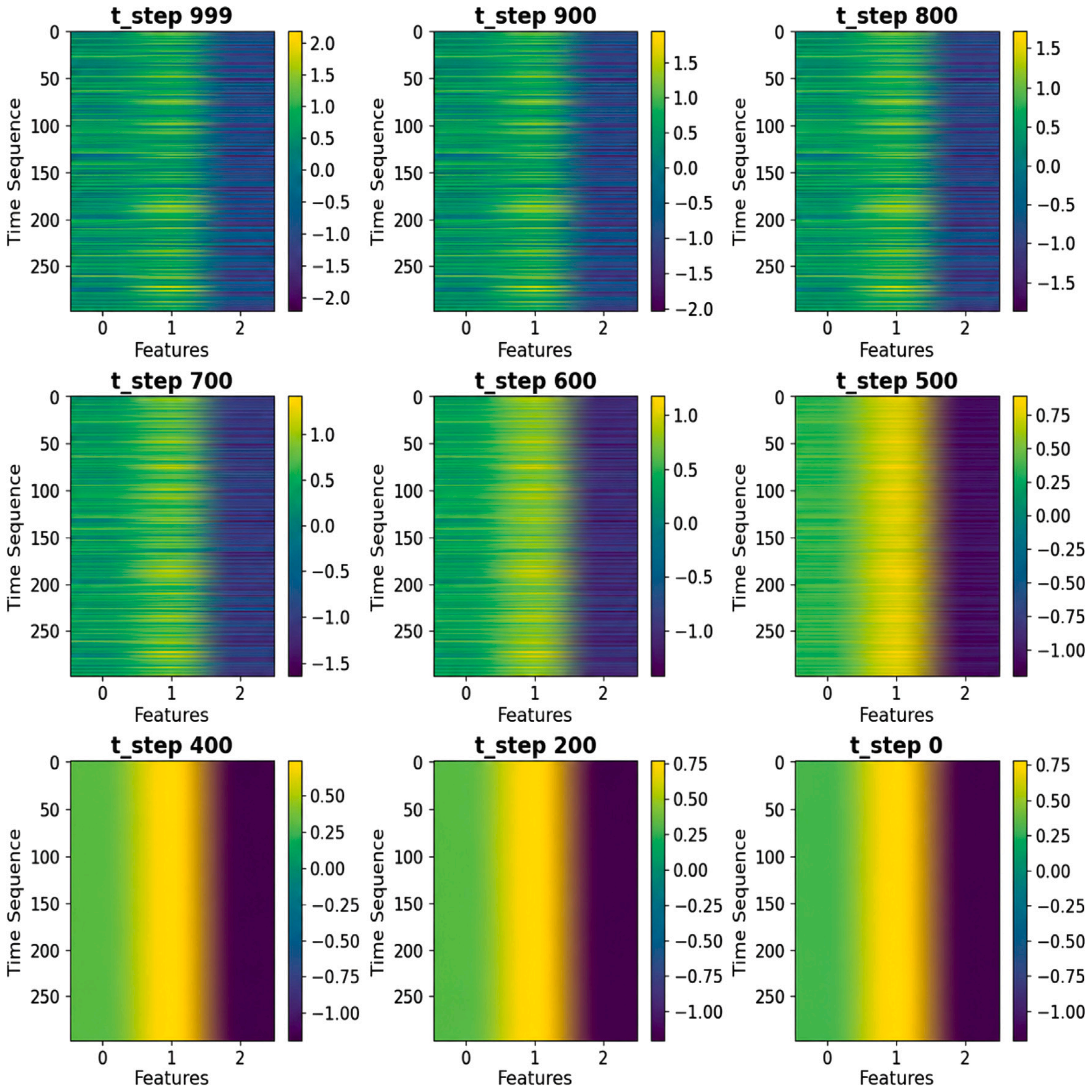

**Fig. 6.** Denoising process impact on time sequences scaling.

**Algorithm 1:** Training Procedure

Sample $x_0 \sim q(x_0)$;
Sample $t \sim \text{uniform}(\{1,\dots,T\})$;
Sample $\epsilon \sim \mathcal{N}(0,\mathbf{I})$;
Compute $x_t = \sqrt{\hat{\alpha}_t}x_0 + \sqrt{1-\hat{\alpha}_t}\epsilon$;
Update $\theta \leftarrow \theta + \eta\nabla_\theta \text{Loss}(\epsilon, \epsilon_\theta(x_t,t))$;

reaching the maximum predefined epochs number, as presented in Algorithm 1.

After the training process, we obtain a model that can identify the noise added to the sample at each $t$ step - $\epsilon_\theta(x_t,t)$. Given that, the noise removal process is described by the following formula [32]:

$$\boldsymbol{x}_{t-1} = \frac{1}{\sqrt{\alpha_t}}\left(\boldsymbol{x}_t - \frac{1-\alpha_t}{\sqrt{1-\hat{\alpha}_t}}\boldsymbol{\varepsilon}_\theta(\boldsymbol{x}_t,t)\right) + \sqrt{\sigma_t}\,\boldsymbol{\varepsilon} \tag{21}$$

With (21), a noised signal $x_t$ is cleaned iteratively from T, the maximal noising level, backward to $t_{back}$ in $T - t_{back}$ iterations, where $t_{back}$ is a hyperparameter tuned for optimal results, as will be described later on.

### 3.3. Data normalization and de-normalization

Using the trained diffusion denoiser as a preprocessing step before activating the heading estimation revealed an unintended side effect: the denoising process distorted the scale of the raw sensor data. This distortion was found to be correlated with the $t_{back}$ parameter, which controls the number of reverse denoising steps. As illustrated in Fig. 6, the data scale progressively decreases with more denoising iterations. Each image shows a full time sequence, where the vertical axis represents time and the horizontal axis displays the $x$, $y$, and $z$ gyroscope channels.

Furthermore, we found that directly feeding denoised data into the heading estimation model led to poor training outcomes, either low accuracy or complete divergence. This behavior was consistent across both synthetic and real-world datasets. To address this, we introduced a three-stage normalization pipeline:

1. Normalize the input before denoising.
2. Apply the diffusion denoiser (Fig. 5).
3. De-normalize the output using pre-recorded mean and standard deviation.

A key design choice in this pipeline is how the data is normalized. We evaluated two alternatives: (i) Sample-by-sample normalization, which normalizes each time sample individually, and (ii) Time sequence normalization, which normalizes the entire sequence using its common statistical properties.

Normalization plays an important role in stabilizing neural-network training by reducing the influence of scale variations and improving numerical conditioning. Sample-by-sample normalization computes normalization statistics independently for each sample. This approach has been shown to improve robustness against scale variations, sensor-dependent gain differences, and distribution shifts between samples, making it attractive for many machine-learning and time-series forecasting applications [44]. Consequently, a sample-by-sample normalization and de-normalization pipeline may improve generalization when the absolute signal magnitude is less important than the local signal pattern. Alternatively, time-sequence normalization computes normalization statistics over the entire measurement sequence. This approach is consistent with normalization principles commonly employed in sequence-learning architectures, where numerical stability must be improved while preserving the temporal relationships within a sequence. For example, layer normalization and recurrent batch normalization were specifically developed to stabilize the training of recurrent neural networks while maintaining meaningful temporal dynamics [45,46]. More recently, time-series learning studies have shown that normalization strategies should reduce scale variability without unnecessarily removing informative temporal characteristics, motivating sequence-level normalization approaches followed by restoration of the original statistics [44,47]. In our implementation, the corresponding de-normalization stage restores the original signal scale after denoising. Since gyrocompassing relies on extremely weak Earth-rotation-induced angular-rate components embedded within long temporal patterns, preserving sequence-level amplitude relationships and temporal correlations may be advantageous. For this reason, both normalization approaches were evaluated experimentally and compared in the following section.

### 3.4. Heading extraction model

We build upon the network architecture described in Section 2.2 (baseline model), and add two additional features to improve training stability and generalization: dropout rate and exponential decay learning rate mechanisms. The dropout is added between the two layers of the bi-directional LSTM model and between the bi-directional LSTM output and the fully connected layer of the model (Fig. 7). Altogether, the heading-extraction model requires training 192,049 parameters.

A preliminary assessment showed that using a dropout ratio of 0.05 provided optimal results. With a learning exponential decay, we gradually reduce the learning rate from the maximum value $\eta_{\max}$ to the minimum value $\eta_{\min}$ over the course of training where the learning rate at epoch $t$ is updated using the exponential decay formula:

$$\eta_t = \eta_{\max} \cdot \gamma^t \tag{22}$$

where $\gamma$ is the decay factor and $t$ is the current epoch number, and the decay factor $\gamma$ is computed as:

$$\gamma = \left(\frac{\eta_{\min}}{\eta_{\max}}\right)^{\frac{1}{N}} \tag{23}$$

This ensures that the learning rate smoothly decays from $\eta_{\max}$ to $\eta_{\min}$ precisely at epoch $N$. In our case, we define $\eta_{\max} = 0.005$ and $\eta_{\min} = 0.0005$. Although those values are small, they were found suitable for our case of highly fluctuating training convergence and low quantity of training data. These parameters are used in conjunction with ADAM optimizer and batch size of 32, as it was found better for training the new system in comparison to the batch size of 100 used by the baseline model.

We adopt the cyclic root mean square error (CRMSE) as the loss function for the heading extraction training. The CRMSE is defined by:

$$\mathcal{L}_{\mathrm{CRMSE}} = \sqrt{\frac{1}{N}\sum_{i=1}^{N}\left(\mathrm{atan}_2\left(\sin(\Delta\psi_i), \cos(\Delta\psi_i)\right)\right)^2} \tag{24}$$

where, $\Delta\psi_i = \psi_{\mathrm{GT},i} - \psi_{\mathrm{pred},i}$, the function $\mathrm{atan}_2$ is applied to compute the shortest angular difference between the ground truth heading $\psi_{\mathrm{GT},i}$ and the predicted heading $\psi_{\mathrm{pred},i}$ for the $i$th sequence, and the CRMSE is then calculated as the root mean square of these differences over the training batch elements ($N$). In the same manner, CRMSE is used as an evaluation metric for the validation and test datasets.

## 4. Results and discussion

This section outlines the evaluation setup and presents the corresponding results. We first describe the datasets used for training and testing, followed by the evaluation metric and the hyperparameter optimization process for the diffusion model. Finally, we report the performance of the complete learning-based gyrocompassing system.

### 4.1. Datasets

Two datasets were used for training and evaluation: (1) Dataset containing real sensor recordings and (2) simulation-based synthetic time sequences. In general, the synthetic dataset was used for training the diffusion model and then the real sensor recording dataset was used to train the heading extraction model and evaluate the overall system performance.

The dataset containing real sensor recordings, was the one used in [27]. This dataset used triaxial gyro readings from Emcore MEMS, model SDC500 IMU[2] as input data, in stationary, leveled, and calibrated conditions, for the model and heading angle from Inertial Labs MRU-P IMU[3] as an accurate source for the heading ground truth (GT). We found the MRU-P IMU an acceptable source for GT as it introduces a static heading accuracy in whole temperature range of $0.2^\circ$ which is below 10% of the accuracy error we expect to obtain from our model. Based on these sensors, 34 time sequences were recorded, each at a fixed heading, where each sequence duration was 100 s, providing a total of 56 min of recorded data. As the raw data obtained from these sensors was first sampled in 600 Hz, each time sequence was down-sampled to 3 Hz for reducing unneeded computational effort and processing time. Then, out of the 34 sequences, 24 time sequences were assigned for training, 4 sequences were assigned for validation, and 6 sequences were assigned for testing. Due to its limited size, the training dataset was augmented by extrapolating each sequence to 100 different heading angles evenly separated within $\pm 20^\circ$ range around the original direction of the time sequence, resulting in a total of 2400 training sequences. This process produced the tensors with the structure [sequences, time,gyro x,y,z]: [2400, 300, 3] for training, [4, 300, 3] for validation, and [6, 300, 3] for testing. Finally, as a last step before being used by the model, the time sequences were flattened from their two dimensions of [300,3] (time as the first dimension and x,y,z gyroscope samples as the second) into a single dimension of [900] and the training dataset was divided into 75 batches of 32 time sequences, yielding a dataset structure of [batches, sequences, flattened gyro data]: [75, 32, 900] tensor for training, [1, 4, 900] for validation, and [1,6, 900] for testing, respectively.

The synthetic dataset was generated based on the simulation of gyroscope readings, in stationary, leveled, and calibrated conditions, obtained by (8), while choosing $\varphi = 0$, dividing the $0^\circ$– $360^\circ$ heading range into $0.5^\circ$ increments, and generating 720 sequences, each with a duration of 100 s. In total we generated a simulative dataset consisting of 1200 min. Eventually, the 720 time sequences were divided following 60-20-20% ratio and, similarly to the process of the real sensor dataset, flattened and divided into 32 size batches providing a tensor structure [batches, sequences, flattened gyro data]: [13, 32, 900], [1, 144, 900], and [1, 144, 900] tensors for training, validation and testing respectively.

[2] SDC500 datasheet: https://www.emcore.com/products/SDC500.

[3] MRU-P datasheet: https://www.inertiallabs.com/mru-datasheet.

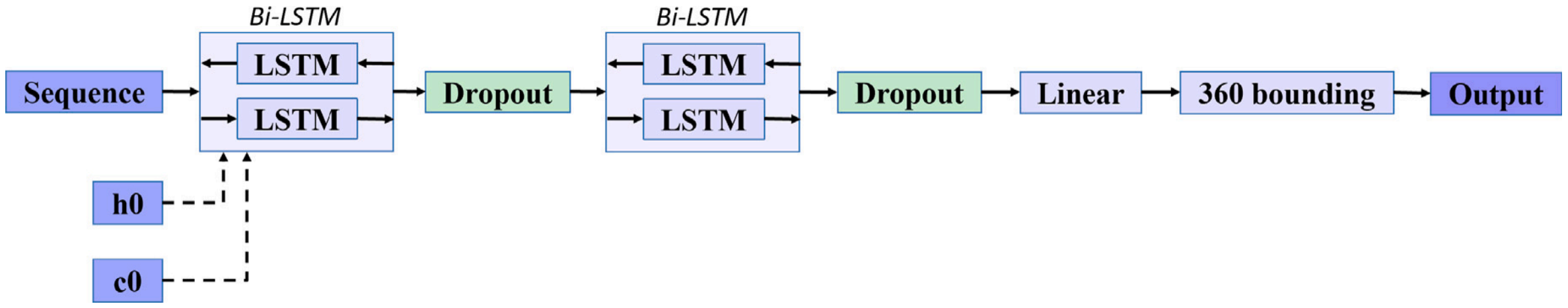

**Fig. 7.** Our proposed heading angle extraction learning model.

**Table 1**
CRMSE results for two normalization approaches applied on the real recorded sensor datasets.

| Normalization method | Training loss [°] | Validation loss [°] |
|---|---|---|
| Time sample | 3.12 | 1.90 |
| Time sequence | 2.93 | 1.39 |

### 4.2. Diffusion model assessment

#### 4.2.1. Normalization approach

The following evaluation examines the performance of the heading extraction model with each one of the two normalization alternatives for the data passing through the diffusion model, as described in Section 3.3 and based on the real sensor dataset.

Table 1 presents the CRMSE result for the training and validation datasets. As can be observed, time sequence normalization is a better alternative as it provides lower loss values for both training and validation datasets.

The observed superiority of time-sequence normalization is consistent with both the characteristics of the gyrocompassing problem and the normalization literature. Sample-by-sample normalization is designed to reduce sensitivity to local scale variations and distribution shifts between samples [44], which can be advantageous when the target information is primarily encoded in local signal patterns. However, in the present application, the heading information is derived from extremely weak Earth-rotation-induced angular-rate components whose relative amplitudes and temporal correlations remain consistent throughout the entire measurement interval. Independently normalizing each time sample may therefore reduce or distort these sequence-level characteristics. In contrast, time-sequence normalization preserves the global statistical relationships of the sequence while still providing numerical stabilization during training, consistent with normalization principles developed for recurrent sequence models [45,46]. Furthermore, recent time-series studies have shown that preserving informative temporal characteristics during normalization can be critical for downstream learning performance [47]. The lower training and validation CRMSE values obtained with time-sequence normalization, therefore, align with both the theoretical expectations and the need to preserve the weak Earth-rotation signature embedded within the noisy gyroscope measurements.

The lower validation error compared to the training error can be attributed to the application of dropout within the stacked bidirectional LSTM layers of the proposed architecture. During training, stochastic dropout independently corrupts the forward and backward recurrent states, disrupting long-range temporal consistency and increasing the training loss. During validation, dropout is disabled and the full bidirectional temporal context is preserved, leading to improved sequence integration and a lower validation error, which is an expected effect of stochastic regularization. Following that, we selected the method of normalizing based on a whole time sequence for the final system configuration.

#### 4.2.2. Optimal denoising steps

In diffusion-based denoising, the parameter $t_{back}$ determines the number of reverse denoising steps. To identify the optimal value, we evaluated the heading extraction model with real sensor data passed through denoising iterations with $t_{back}$ values ranging from 980 down to 100.

As shown in Fig. 8, the best performance with the validation set was achieved at the upper $t_{back}$ range of 900–980, yielding a loss of 1.7–1.9° compared to losses above 2.5° for the other $t_{back}$ values.

Then, to discriminate and down select between the possible optimal $t_{back}$ values, we also examined the training behavior for various values of $t_{back}$, as Fig. 9 illustrates. There, as training continues to improve or maintain its performance over epochs, as $t_{back}$ reduces, test performance deteriorates relative to training (Figs. 9(a), and 9(b)). This behavior is indicative of overfitting, an unwanted result. In contrast, in Fig. 9(c), the CRMSE loss exhibits consistent and convergent behavior across training, validation, and testing sets, suggesting improved generalization and stability. Therefore, to combine performance and generalization, we worked with $t_{back} = 950$ in the final system configuration.

### 4.3. Gyrocompassing evaluation

Based on the results presented in the previous section, we set $t_{back} = 950$ and applied time-sequence-based normalization to the data. We then trained the heading extraction model, integrated with a pre-trained diffusion model, on the real sensor dataset corresponding to a 100-second gyrocompassing duration. The performance of our proposed system (Our diffusion denoiser with the improved baseline gyrocompassing) was evaluated on test data and compared with five benchmarks:

1. A classical model-based gyrocompassing approach as described in Section 2.1.
2. The baseline deep learning-based gyrocompassing as described in Section 2.2.
3. Our improved baseline gyrocompassing (without denoising).
4. Our diffusion denoiser with the baseline deep learning-based gyrocompassing.
5. A wavelet low-pass filter with our improved baseline gyrocompassing.

The classical model-based gyrocompassing was evaluated over durations ranging from 10 to 100 s, and all models were trained on the same dataset. The wavelet low-pass filter was applied using a level-4 discrete wavelet decomposition with a Symlet-4 (sym4) mother wavelet. High-frequency noise components were attenuated through soft thresholding of the detail coefficients, resulting in improved signal-to-noise characteristics without temporal distortion, consistent with established wavelet theory and denoising methodologies [48–52].

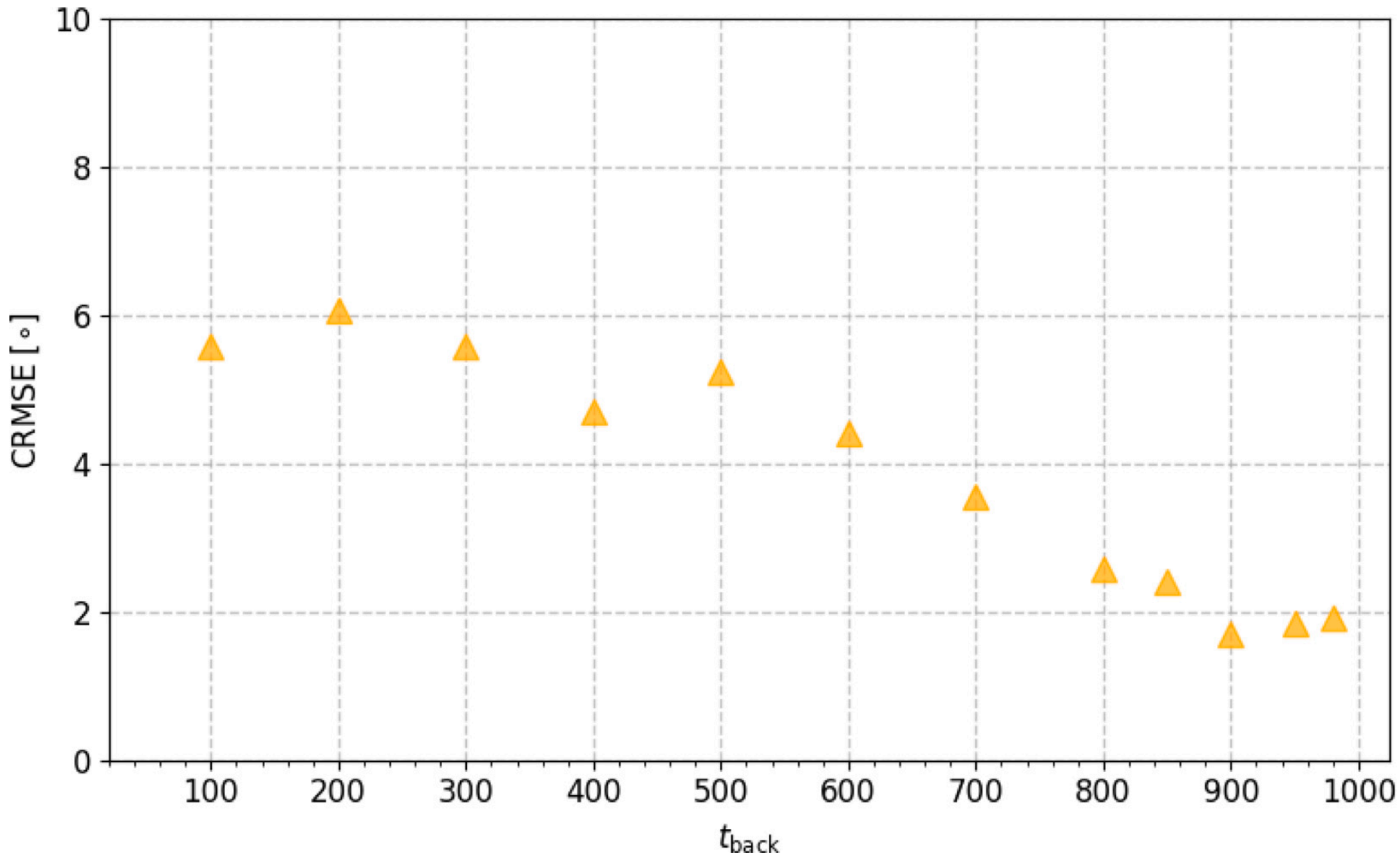

**Fig. 8.** CRMSE results as a function of the $t_{\text{back}}$ parameter.

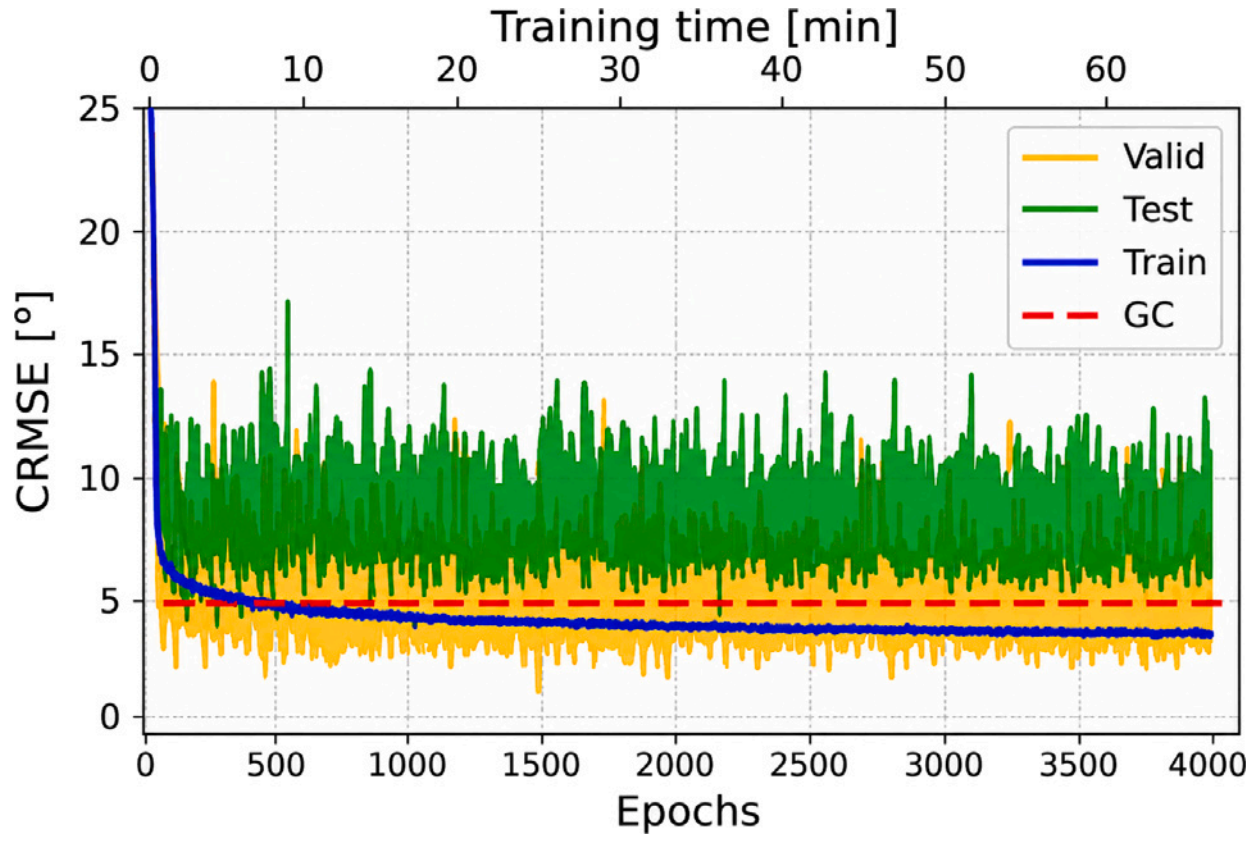

(a) $t_{\text{back}} = 700$

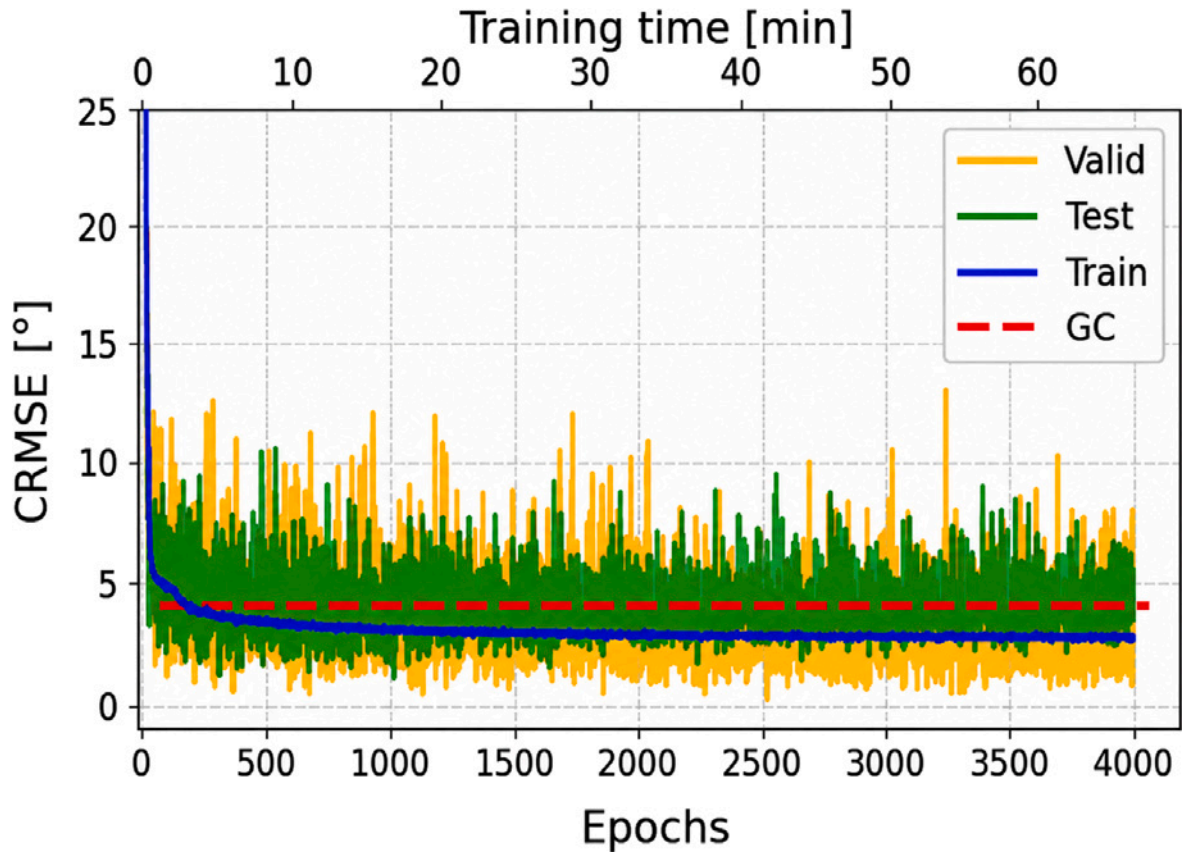

(b) $t_{\text{back}} = 900$

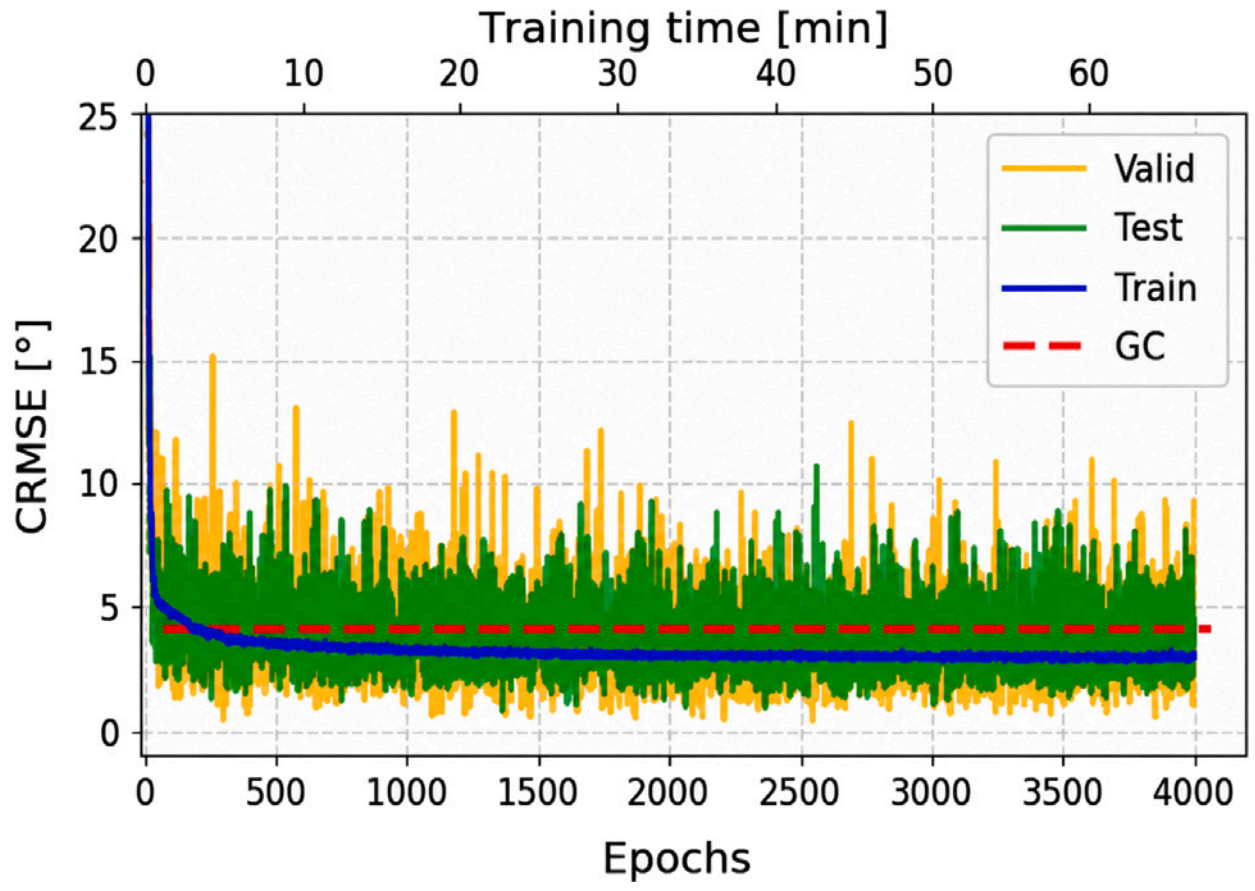

(c) $t_{\text{back}} = 950$

**Fig. 9.** Overfitting in heading extraction model training with different $t_{back}$ values.

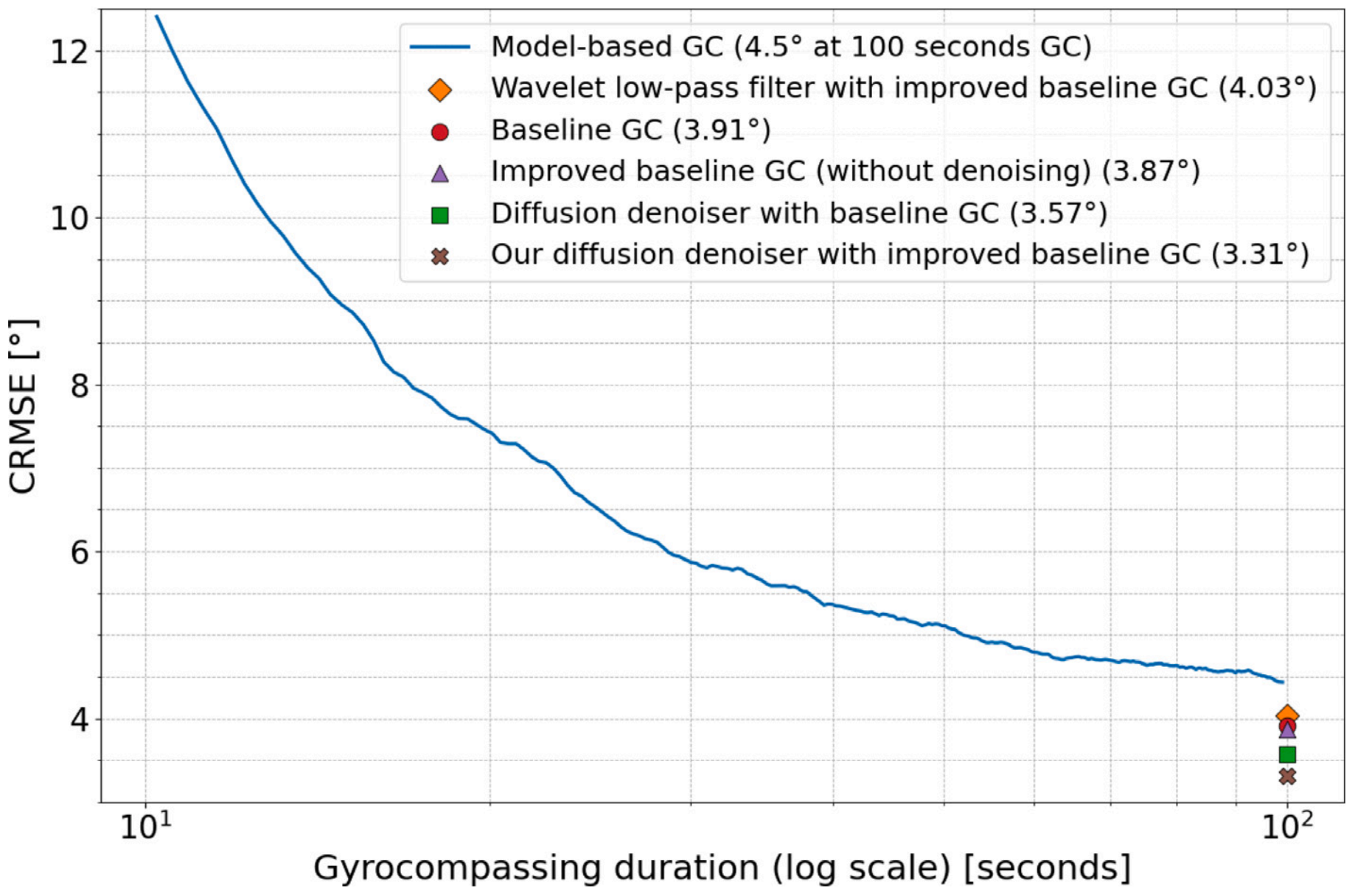

**Fig. 10.** Comparison of CRMSE versus gyrocompassing duration for the evaluated methods, including the proposed system, the classical model-based approach, the baseline deep learning-based approach, the improved baseline (without denoising), diffusion denoising combined with the baseline deep learning-based approach, and wavelet low-pass filtering combined with the improved baseline.

Fig. 10 illustrates the accuracy for each approach relative to the duration of the model-based gyrocompassing. Table 2 presents the CRMSE results after 100 s of gyrocompassing for the six approaches. Our method consistently outperformed all benchmarks, achieving the largest gain over the classical model-based approach (26%) and smaller but consistent gains over learning-based and denoising-based variants (7%–18%). The results also provide an ablation assessment of the main design choices considered in this work. Comparing the baseline deep learning-based gyrocompassing (3.91 ° CRMSE) with the improved baseline gyrocompassing (3.87 ° CRMSE) isolates the contribution of the heading-extraction model modifications, namely the introduction of dropout and exponential learning-rate decay, showing a modest improvement of approximately 1%. In contrast, combining the diffusion denoiser with the baseline heading extractor reduces the CRMSE to 3.57 °, corresponding to an improvement of approximately 9% relative to the baseline and indicating that the diffusion-denoising stage contributes more substantially than the heading-extraction modifications alone. The complete system achieves the lowest CRMSE of 3.31 °, demonstrating that the denoising and heading-extraction improvements are complementary. Finally, comparison with the wavelet-based denoising alternative (4.03 ° CRMSE) shows that the observed performance gain is not solely due to the addition of a denoising stage, but also to the diffusion model's ability to better preserve the weak Earth-rotation-induced signal while suppressing sensor noise. These results suggest that diffusion-based denoising constitutes the dominant contributor to the overall performance improvement, while the heading-extraction model modifications provide an additional refinement.

**Table 2**
CRMSE values and relative improvement rates for the evaluated methods, including the proposed system, the classical model-based approach, the baseline deep learning-based approach, the improved baseline (without denoising), diffusion denoising combined with the baseline deep learning-based approach, and wavelet low-pass filtering combined with the improved baseline. Improvement indicates the percentage reduction in CRMSE achieved by the proposed system relative to each benchmark.

| Approach | CRMSE [°] | Improvement [%] |
|---|---|---|
| Classical model-based gyrocompassing | 4.50 | 26% |
| Wavelet low-pass filter with our improved baseline gyrocompassing | 4.03 | 18% |
| Baseline deep learning-based gyrocompassing | 3.91 | 15% |
| Our improved baseline gyrocompassing (without denoising) | 3.87 | 14% |
| Our diffusion denoiser with the baseline deep learning-based gyrocompassing | 3.57 | 7% |
| Our proposed system | 3.31 | – |

## 5. Conclusions

Achieving timely and accurate gyrocompassing using low-cost gyroscopes without external navigation aids remains a significant challenge. To address this, we propose a gyrocompassing framework that integrates a diffusion-based denoising model, trained using physics-based synthetic inertial data, and designed to recover extremely low-amplitude inertial signals, with an enhanced heading extraction model. The diffusion model is implemented as a five-layer bidirectional LSTM network, while the heading extraction builds upon a dual-layer bidirectional LSTM, augmented with dropout and exponential learning rate decay for improved generalization. Inputs are processed using a customized normalization and denormalization pipeline designed to minimize heading estimation error. The training strategy follows a two-stage process: First, the diffusion model is trained on a synthetic, simulation-based dataset, and then, having the diffusion model's weights frozen, the heading extraction model is trained using real sensor recordings. This separation enables diffusion-model training using synthetic data, followed by sensor-specific optimization of heading estimation, thereby improving robustness and generalization.

Experimental results demonstrate that over a 100-second gyrocompassing duration, the proposed system achieves a 26% reduction in CRMSE compared to the classical model-based method and a 15% reduction compared to a baseline data-driven approach. The ablation-style evaluation, further shows that the diffusion-denoising stage constitutes the primary source of performance improvement, producing a substantially larger reduction in CRMSE than the heading-extraction model modifications alone. The modified heading-extraction model contributes additional gains through improved training stability and generalization, while the combination of both components yields the best overall performance. The evaluation of the wavelet-based denoising alternative further demonstrates that the improvement originates not merely from applying a denoising stage, but specifically from the diffusion model's ability to recover weak Earth-rotation-induced information embedded within the sensor noise. These findings confirm that the diffusion model effectively mitigates gyroscope noise, while the improved LSTM-based heading extractor delivers additional accuracy improvements within the same gyrocompassing duration.

These results suggest that the incorporation of diffusion models into inertial sensor signal processing, represents a promising direction for robust heading estimation, particularly in noisy environments.

Overall, the proposed approach has the potential to broaden the use of standalone gyrocompassing, particularly with small SWaP-constrained, environmentally stabilized, IMU-based navigation systems, commonly found in autonomous vehicles. While MEMS-based IMUs are known to be sensitive to environmental conditions such as temperature variations and mechanical vibrations, the gyrocompassing scenario addressed in this work assumes an environmentally stabilized configuration, as commonly adopted in practice. In this setting, environmental disturbances can be sufficiently mitigated, allowing MEMS sensors to provide measurements of adequate quality for reliable Earth-rotation-based heading estimation. More broadly, the integration of diffusion-based denoising may facilitate the adoption of MEMS-based IMUs in a wider range of applications where high-fidelity inertial sensing is required. Future research directions include evaluating the proposed framework with data obtained from dynamic environmental conditions and additional sensor modalities, investigating denoiser and heading extraction models adaptation strategies tailored to the changing noise and bias characteristics with environmental conditions and the specific noise characteristics of individual sensor models (potentially trading off generalization for improved accuracy), and exploring the replacement of the Bi-LSTM backbone with diffusion-driven autoencoder architectures, as well as alternative sequence-modeling frameworks such as Transformer-based and TCN architectures, and self-supervised inertial learning approaches, as alternative denoising and representation-learning paradigms, to further enhance denoising efficiency, and overall gyrocompassing performance, and to evaluate the relative suitability of these architectures for recovering weak Earth-rotation-induced signals in low-cost MEMS-based gyrocompassing applications.

## CRediT authorship contribution statement

**Gershon Ben-Arie:** Writing – original draft, Visualization, Validation, Software, Resources, Project administration, Methodology, Investigation, Formal analysis, Data curation, Conceptualization. **Daniel Engelsman:** Writing – review & editing. **Rotem Dror:** Writing – review & editing. **Itzik Klein:** Writing – review & editing, Supervision.

## Declaration of competing interest

The authors declare that they have no known competing financial interests or personal relationships that could have appeared to influence the work reported in this paper.

## Data availability

Data will be made available on request.